# The Tangent Search Algorithm for Solving Optimization Problems

Abdesslem Layeb

Computer science and its application department, NTIC faculty, university of Constantine 2

Abdesslem.layeb@univ-constantine2.dz

**Abstract :**

This article proposes a new population-based optimization algorithm called the Tangent Search Algorithm (TSA) to solve optimization problems. The TSA uses a mathematical model based on the tangent function to move a given solution toward a better solution. The tangent flight function has the advantage to balance between the exploitation and the exploration search. Moreover, a novel escape procedure is used to avoid to be trapped in local minima. Besides, an adaptive variable step size is also integrated in this algorithm to enhance the convergence capacity. The performance of TSA is assessed in three classes of tests: classical tests, CEC benchmarks, and engineering optimization problems. Moreover, several studies and metrics have been used to observe the behavior of the proposed TSA. The experimental results show that TSA algorithm is capable to provide very promising and competitive results on most benchmark functions thanks to better balance between exploration and exploitation of the search space. The main characteristics of this new optimization algorithm is its simplicity and efficiency and it requires only a small number of user-defined parameters.

1. Introduction

Optimization is an important field used to satisfy the growing needs of the economic and industrial sectors in terms of maximization of performance or minimization of costs. We find also the optimization in many research areas such in medicine, biology, chemistry, etc. The considerable development in algorithmic and computing power makes the optimization field more practicable to solve real optimization problems. Mathematically, optimization consists to find the best values of decision variables of an objective function that make its value maximal or Minimal.

To solve the optimization problems, several methods were developed that can be classified mainly into two classes: Exact methods (complete methods) and approximate methods (approach methods). Unfortunately, the use of exact methods for solving many optimization problems is impracticable for several reasons; many real optimization problems have great computational cost, non-convex domain search, and many local optima [1,2,3]. Therefore, the approximate methods constitute a better alternative to deal with complex optimization problems. Among the well-known approximate methods, we can cite the heuristics, metaheuristics and the approximate numerical methods [4,5].

Metaheuristics constitute a class of methods used to solve optimization problems. Globally, they provide good quality solutions in reasonable time but unfortunately, they do not guarantee the optimality. these methods use a high level of abstraction, allowing them to be suited to a wide range of optimization problems. Metaheuristics are generally iterative stochastic algorithms, which progress toward an optimum through series of transformations by evaluating an objective function. The increasing interest in metaheuristics is fully justified by the limitation of gradient based methods and the development of machines with big computational capacities, which has helped to design increasingly complex metaheuristics to deal with NP-hard problems [6].

Actually, many metaheuristics have been developed ranging from simple local search to more complex global search algorithms that can be divided into two categories: single solution based metaheuristics and population based metaheuristics. Single solution methods or local search methods, also called neighborhood search methods, start from an initial solution and, by successive refinements, build sequences of solutions of decreasing costs for a minimization problem. Local methods are among the first approximate algorithms used for continuous algorithms. They are divided on two classes gradient search based algorithms such as gradient descend algorithm, newton Raphson algorithm [7], and gradient free search algorithms which do not require the function derivations such as Nelder-Mead simplex algorithm [8], Pattern Search (PS) [9], Simulated Annealing (SA) [10], random Hill Climbing (HC) [11]. Although, these methods are simple and find a solution in reasonable time, they suffer from the local minimum and they depend on the initial solution. That's why the scientists investigated in global search methods. Unlike local search methods, global methods aim to reach one or more global optima. With the exception of simulated annealing, Tabu Search [12] and variable neighborhood search algorithms [13], the most of global optimization algorithms are population based metaheuristics. They modify and improve a population of candidate solutions. Generally, the population based methods englobe two phases: intensification and exploration in order to escapee from the local minima. Among the most popular global search algorithms Genetic Algorithms (GA) [14], Differential Evolution algorithm (DE) [15], Particle Swarm Optimization (PSO) [16], Artificial Bee Colony (ABC) [17], Ant Colony optimization (ACO) [18], etc.

Besides, the metaheuristics are generally inspired from several metaphors like nature, physics, chemistry, animal behaviors. For example, SA was inspired by the thermodynamic process of metal annealing, GA was inspired by the process of natural evolution, PSO algorithm was inspired by the behavior of bird's swarm. Firefly algorithm (FA) was inspired by the bioluminescent communication behavior of fireflies [19]. ACO algorithm was inspired by the foraging behaviors of ant colony. Chemical reaction optimization (CRO) algorithm was inspired by the chemical reactions [20]. Gravitational Search Algorithm (GSA) was inspired by the gravity force in physics [21]. In these nature-nspired algorithms, the natural motions of swarm and objects are modeled by mathematical equations used to guide the solutions towards the best region of the search space. For an in-depth review of nature-inspired algorithms for optimization, we can refer to [22].

Recently, mathematical inspired metaheuristics have immerged. These metaheuristics use geometric, arithmetic, or analytic functions to guide the search process. For example, a metaheuristic based on sine cosine function is proposed in [23]. Metaheuristics based on spherical search are proposed in [24]. On [26], the authors present new metaheuristics based on arithmetic operations. We find also several metaheuristics based on fractal geometry [27]. All the previous mathematical inspired metaheuristics have proved their effectiveness in many hard optimization problems.

Unfortunately, according to the theorem of no free lunch, no algorithm can offer better performance than all the others on all the optimization problems. That's why, more new metaheuristics are investigated each to develop a performant global optimization method. In this paper a novel mathematical inspired optimization algorithm based on the mathematical tangent function called Tangent Search Algorithm (TSA), is presented. For better balance between exploration, and intensification, TSA is composed of three mains components, exploration search component, which ensures a good exploration, intensification search components which guarantees more intensification search around the best solution, and to escape from local minima, an escape procedure is added. The main characteristics of TSA algorithm is its simplicity, one simple mathematical equation guides the optimization process and there is no complicated procedures or structures to deal with hard tests. Besides, a variable step-size is also used to well control the magnitude of the movement in the optimization process.

TSA is tested on well well-known benchmarks: some popular classical test functions, five most challenging tests, the CEC 2017 suite, the CEC 2020 suite, and four constraint design problems. Quantitative and qualitative study were done to validate the effectiveness of TSA algorithm. The experimental results show that TSA algorithm is capable to provide very promising and competitive results on most benchmark functions thanks to better balance between exploration and exploitation of the search space.

The reminder of the paper is organized as follows. The following section presents in detail the proposed TSA algorithm. Section 3 contains the experimental results and discussion. Finally, Section 4 concludes the work.

## 2. Algorithm description
### 2.1. Motivation of the work

The main motivation of this work is to present a simple and an effective optimization algorithm. Indeed, the Tangent Search Algorithm is based on a simple mathematical function which is the tangent function. This function offers a great capacity to well explore the search space, the variation of this function between -∞ et +∞, and the periodicity of this function help to maintain a good balance between exploration and intensification. In TSA algorithm, all the motion equation are governed by a global step with the form "step*tan(θ)", where the tangent function plays a role of flight function like in levy flight function, so we call it tangent flight for more simplicity. In addition, the NFL theorem says that there is no algorithm which can deal with all optimization problems. Therefore, there are still optimization problems that that can solved by newer algorithms.

### 2.2. Mathematical model of the algorithm

The most optimization algorithm whether derivative based or derivative free are based on following like descend equation:

$X^{t+1} = X^t + step*d$ **Eq.1**

Where step is the size of move and d is the direction of the move, the difference between them lies in how to compute the step. The derivative based methods use Gradient or Hessian information to compute this step, while in free derivative methods, like metaheuristics, they use stochastics step to converge to global optima. For example, genetic algorithm uses

Gaussian mutation as step, in differential evolution, the step size is computed by using the difference between individual of the current population, in Cuckoo Search (CS) the step size is computed by using a levy flight function [25], etc.

A good algorithm is the one that has a good step size, a great value will favorite the exploration while a small value will favorite the exploitation. In this work, we propose a new step size based on the tangent function that we call tangent flight. In the figure 1, we see the behavior of tangent flight over different angle' ranges. The great distribution of the points is around the angle 0, which means that it favorites the exploitation, while the distant and the isolated points favorite the exploration of the search space. To control this high magnitude of exploration and to convergence quickly, the tangent flight is multiplied by a decreasing function, the figure 2 shows the behavior of the tangent flight multiplied by a decreasing logarithm based function. In this figure, the exploration search is large at the early iterations and diminished at the last iterations.

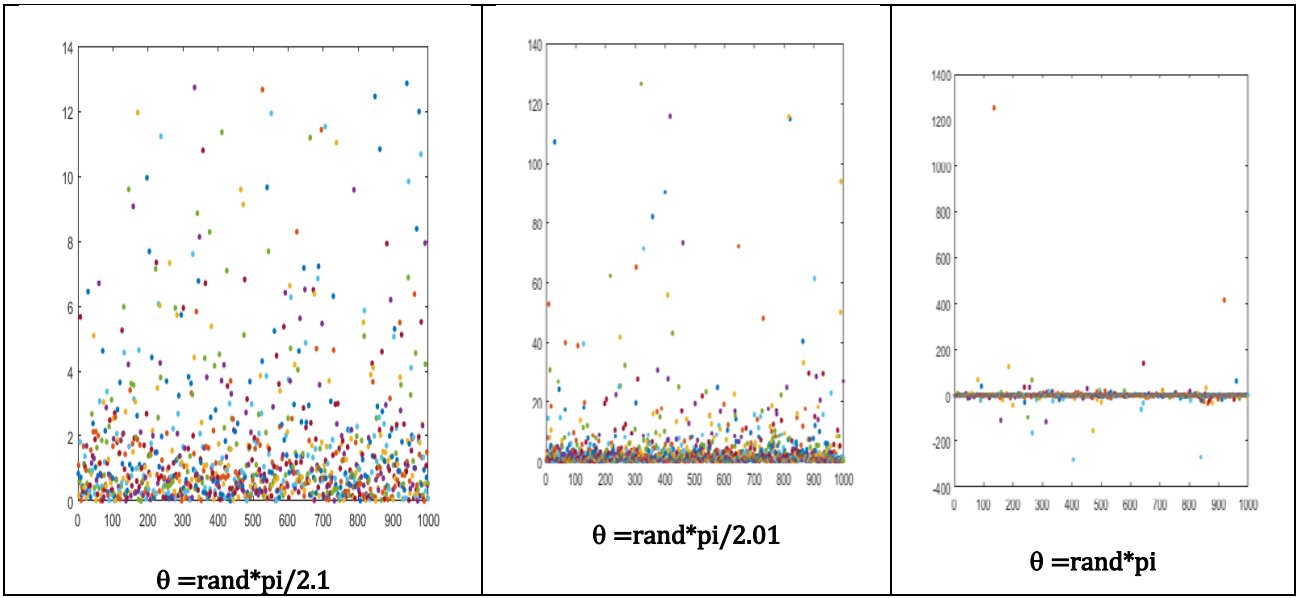

**Figure 1.** The behavior of tangent function with different range angle

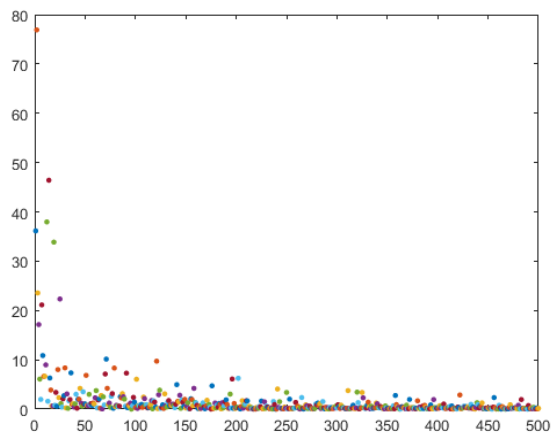

**Figure 2.** The behavior of tangent function with decreasing step

### 2.3. Modelling the algorithm

A successful optimization algorithm should have a better balance between intensification and exploration, too intensification makes the algorithms converge quickly to a local minimum, and too exploration makes the algorithm too slow and sometimes it diverges. To reach this goal, TSA is composed of three main components: intensification, exploration, and escape local minima components. The purpose of the exploration phase is to well explore the search space and to find the most promising candidates. While the intensification component is used to direct the search process towards the best current solution in the population. Finally, the escape local minima procedure is applied at each iteration on a random search agent (solution) in order to avoid to be trapped in a local minimum. The figure 3 shows the flowchart of the proposed population based metaheuristics.

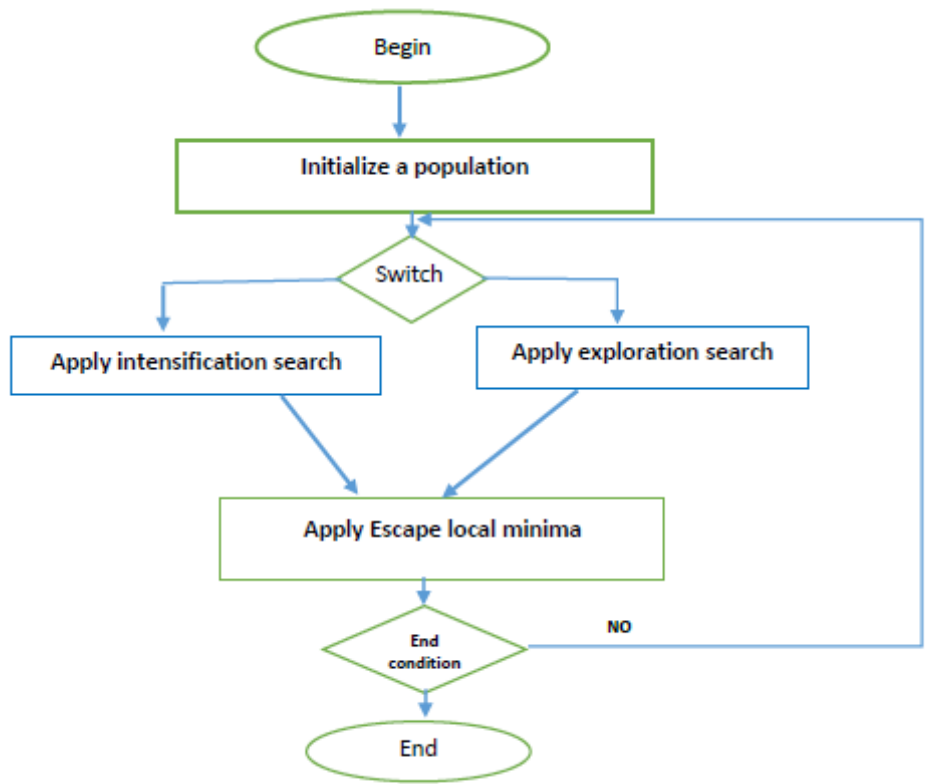

**Figure 3.** The flowchart of TSA

Now, we explain in more details the outlines the TSA algorithm.

*2.3.1 Initialization of the initial population*
Like other population-based optimization algorithms, TSA starts by generating a random initial population within the boundaries of the solution space. The initial solution is uniformly distributed over the search space and it's computed by the following equation.

X0= lb+(ub-lb).*rand(D);                **Eq.2**

Where lb, ub are the lower and upper bounds of the problem, rand function generates uniformly distributed random numbers in the range [0, 1], and D is the dimension of the problem.

*2.3.2 Intensification search*

In the intensification search, TSA makes, first, a random local walk guided by the following equation ( eq.3), then, some variables of the obtained solution is replaced by the values of the corresponding variable in the current optimal solution by using the equation 4. The proportion of the replaced variables is equal to 20% for a problem with dimension higher than 4, and 50% for problems having less or equal to 4 variables.

$$X_i^{t+1} = X_i^t + step * \tan(\theta) * (X_i^t - optS_i^t)$$   **Eq.3**

$$X_i^{t+1} = optS_i^t \quad \text{if variable i is selected}$$   **Eq.4**

Consequently, the obtained solution $X_i^{t+1}$ has a proportion of similarity under 50% with the best current solution which helps to enhance locally the current solution.

Each obtained solution X is repaired by the following equation if its values overflow the problem bounds lb and ub.

X(X< lb) =  rand*(ub - lb) + lb;

X(X> ub) =  rand*(ub - lb) + lb;    **Eq.5**

*2.2.3 Exploration search*

Contrary to local search methods, the metaheuristic based population have great capacity of exploration thanks to the global random walk. In this algorithm, TSA uses a product of variable step size and the tangent flight to make global random walk thanks. The tangent function helps to explore efficiently the search space. indeed, θ near to pi/2 will make the tangent value bigger and the obtained solution will be far from the current solution, and θ near to 0 gives small values to the tangent function and the obtained solution will be near to the current solution. So, the following equation of the exploration search merges between the global and local random walk. The exploration search equation is applied on each variable with a probability equal to  1/D, where D is the dimension of the problem.

$$X_i^{t+1} = X_i^t + step * \tan(\theta)$$    **Eq.6**

The intensification search and the exploration search are applied according a given probability called Pswitch as described by the pseudo code in the figure 4.

```
while t <=MAX_FE  // the maximum of function evaluations

For each agent search Xi (i=1:T, T:size of the population)

If  rand < Pswitch

Apply Intensification search;

Else

Apply exploration search;

end

End for each
If  rand<Pesc
Select randomly an agent search (G);
Apply escape local  minima (G);

t=t+1;
End while
```
**Figure 4.** the pseudo code of TSA

*2.3.4 Escape local minima procedure*

To escape from the local minima stagnation problem. TSA incorporates a mechanism to deal with this problem by using a specific procedure as displayed in the following figure. The procedure has two part executed with some probability *Pesc*. In each iteration, one agent search is selected randomly, then one of the following equations is executed. Moreover, new random solution can replace worst solution with probability 0.01.

X =X + R.*(optS- rand*(optS-X)) ;    **Eq.7**

X = X+ tan(teta)*(ub-lb);    **Eq.8**

```
Input solution X
generate step R=10*sign/log(1+t);
if rand <=0.99
if rand<0.8
    X =X + R.*(OPTX- rand*(OPTX-X)) ;
else
  X = X+ tan(teta)*(ub-lb);
```

```
End
Else
Replace X by new random solution generated by the eq.1
end
Output new solution X
```

**Figure 5.** The pseudo code of Escape local procedure

*2.3.5 Parameters explanation*

Now, we explain the parameters used in TSA. TSA uses few parameters compared to many optimization algorithms, *Pswitch, Pesc, step, θ* are the main parameters used to emphasis the exploitation and exploration search. The balance between local and global random walks is controlled by the switching parameter *Pswitch* ∈ [0,1], the second parameter is Pesc ∈ [0,1] which is the probability of escape procedure. The parameter step plays a role of the step-size parameter in descend methods, it is used to direct and emphasis the exploitation and the exploration of the search process. TSA uses a variable step-size in order to well converge to the best solution and avoid lack of precision. At early stage, TSA adopts a large step-size, with the progression of the search process, the step-size decreases nonlinearly from iteration to iteration. This adaptive-size behavior helps TSA to achieve a good balance between exploration and exploitation. Beside the tangent flight has a great influence on the step size, it gives it oscillatory and periodic behavior. To adapt the exploitation and the intensification search process, TSA uses a nonlinear decreasing scheme for the adaptive step-size based on the logarithm function. The logarithm function is slow function which helps to maintain a fine convergence. On another hand, we have noted that the use of different step-size functions in the same algorithm gives better results especially for functions with hard convergence. Therefore, for more efficiently, TSA uses two variants of the step-size. the first step-size variant is used in the intensification search and it is computed as follows:

step1 = 10*sign(rand-0.5)*norm(optS)*log(1+10*dim/t )          **Eq.9**

and in the exploration search, the step-size is computed as follows:

step2 =  1*sign(rand-0.5)*norm(optS-X)/log(20+t)          **Eq.10**

Where: norm() is a given mathematical norm, in following experiment we have used the Euclidian norm, however TSA could use other norms. X is the current solution and optS is the best current solution used to guide the search process towards the best solution. By consequence, the step-size decreases by decreasing the value of the current iteration *t*, this variable step-size guarantees a well convergence by reducing the search speed in last iterations. The component *sign (−,+)* controls the direction of exploration and exploitation search.

Finally, the angle θ plays too a great role in the convergence of TSA algorithm. a value θ =pi/2 will make the algorithm diverges. After an empiric study, θ is chosen randomly in the range of [0, pi/2.1] in the intensification search and in the range [0, pi/3] in the exploration search. Consequently, the values of tangent function are comprised between 0 and 13.34 (tan(pi/2.1).

*Comments on parameters*

The parameter tuning of approximate optimization algorithms is an important phase to obtain an algorithm of good quality. Worst parameters can deliver worst solutions, and too parameters will disturb the performance of the algorithm. TSA algorithm has few parameters to be set by the users, the Pswitch parameter to switch between the intensification and the exploration, the Pesc the probability of the escape procedure, the number of iterations t, and the number of search agents in the initial population. The sensitivity investigation of these parameters has been discussed by varying their values. After empiric study, we have found that STO doesn't need a large size of the population, a value between 20 and 30 gives good results. On the other hand, the stepsize step is adjusted directed by the algorithm.

### 2.4. Computational complexity

In this section, the computational complexity of TSA is discussed. The complexity of TSA is as follows:

O(TSA)=O(generation initial population)+ O( population search phase)

The complexity of the generation of the initial population is equal to O(T*D), where D is the dimension of the problem and T is the size of the population. The complexity of one iteration in the worst cases is equal to:

T*(O(norm function) + O( test function evaluation)+ O(motion equation computation))

Thus, the global computational complexity of TSA is equal to:

O(MAX_FE* T*(D+ O( test function evaluation)).

Where MAX_FE is the maximum function evaluations, T is the size of the population and D is the dimension of the problem. So, the complexity depends globally on complexity of the objective functions. If we assume that the complexity of the objective function is equal to O(D), then the whole complexity will be equal to O(MAX_FE*T*D), and it is largely inferior to many metaheuristics complexity.

**3. Experiments and discussion**

The proposed algorithm is implemented under MATLAB R2016a environment, and all experiments were carried out on a Windows 10 64-bit computer with an Intel i3 (2.3 GHz) processor and 4 GB RAM. To assess the optimization performance of TSA algorithm, a comprehensive comparison experiment is conducted on four suites of benchmark functions. The first one is composed of well-known classical functions [28], the purpose is to study the behavior of TSA. The second set is composed of five tests are known to be hard for any optimizer [29], the goal is to demonstrate the effectiveness of TSA on hard test functions. The third benchmark is CEC2017 [30] where the goal is to study the behavior of TSA on shifted, rotated and composite functions. Besides, the CEC2020 [31], composed of different hard tests taken from previous CEC competitions, is used to analysis the sensitivity and the importance of the tangent flight mechanism in the effectiveness of TSA algorithm. Finally, the last benchmark is a set of four design engineering problems with constraints. Moreover, TSA is compared to well-known and powerful optimization algorithms.

Finally, to validate statistically the results found, the Wilcoxon signed rank and Kruskal-Wallis tests are used. The Wilcoxon test is one of the well-known non-parametric statistical tests used for comparing the mean ranks of different data samples in a two by two manner. However, Kruskal-Wallis test is used for comparing the mean of the metaheuristics' ranks which helps to evaluate the whole rankings of different metaheuristic algorithms. The Kruskal-Wallis test is used to compare samples when the distribution does not prove to be normal which is our case. Indeed, the one-sample Kolmogorov-Smirnov test attests that the results of the algorithms do not come from a normal distribution, so the appropriate multi-comparison test is the Kruskal-Wallis test. To implement these tests, the normalized score of each algorithm per each problem is computed as a number between [0,1]. All the above statistical tests are computed with a level of significance α=0.05. In the following Kruskal-Wallis tests, the red lines show the algorithms with significant difference and rejection of the null hypothesis, while, the gray lines indicate the algorithms without significant difference.

**3.1 Convergence analysis**

The first experiment was performed on 2D variables of some classical test functions; the number of function evaluations is set to 2000, the number of search agents is fixed to 20. The goal of this experiment is to study the behavior of TSA, the capacities of the intensification and exploitation search, and if it can avoid local minima in multimodal functions. As displayed in Table 1 and Figure. 6, unimodal test functions (Sumsquare and Rosenbrock functions) which have one global optima solution, and multi-modal (Crossintray, Ackley, Shubert, Rastrigin, Foxholes, Schaffer and Griewank functions) which have many local optima, are used in this experiment.

The history search, the best and the mean convergence curves of TSA are also shown in Figure. 6. As we can see, the convergence curve of the best solutions shows the exploration and the intensification capacity of TSA. The results confirm that TSA algorithm preserves a good balance between exploration and exploitation to find the best solution and avoid the local minima. The mean convergence curve has the same behavior of the best convergence curve in many tests which confirms that most of the search agents move toward the good region of the space search. On other hand, the search history metric is used to determine how TSA explores and exploits in a given search space (figure. 6). It is observed from the figure that TSA has the ability to explore the most promising area in the given search space of the test functions. For unimodal test functions, the intensification search capacity of the TSA guides the solution points to the promising area. In multimodal test functions the points sample are distributed over the local minima, but the great distribution is around the best global minimum. Therefore, TSA is not trapped in local minima and it explores well the promising search space thanks to the intensification, exploration and the escape local procedure search.

**Table 1.** Classical test Functions [28]

| Test function | Name | Type | D | Range | Optimum |
|---|---|---|---|---|---|
| $f_{c01}(x)=\sum_{i=1}^{D} x_i^2$ | Sphere | US | 30 | $[-100,100]$ | 0 |
| $f_{c02}(x)=\sum_{i=1}^{D}|x_i|+\prod_{i=1}^{D}|x_i|$ | Schwefel 2.22 | UN | 30 | $[-10,10]$ | 0 |
| $f_{c03}(x)=\sum_{i=1}^{D}\left(\sum_{j=1}^{D} x_i\right)^2$ | Schwefel 1.2 | UN | 30 | $[-100,100]$ | 0 |
| $f_{c04}(x)=\max_i\{|x_i|, 1\le i\le D\}$ | Schwefel 2.21 | US | 30 | $[-100,100]$ | 0 |
| $f_{c05}(x)=\sum_{i=1}^{D} 100(x_{i+1}^2-x_i^2)^2+(x_i-1)^2$ | Rosenbrock | UN | 30 | $[-30,30]$ | 0 |
| $f_{c06}(x)=\sum_{i=1}^{D}([x_i+0.5])^2$ | Step | US | 30 | $[-100,100]$ | 0 |
| $f_{c07}(x)=\sum_{i=1}^{D} ix_i^4+random[0,1)$ | Quartic | US | 30 | $[-1.28, 1.28]$ | 0 |
| $f_{c08}(x)=\sum_{i=1}^{D}(x_i^2-10\cos(2\pi x_i)+10)$ | Rastrigin | MS | 30 | $[-5.12, 5.12]$ | 0 |
| $f_{c09}(x)=20+e-20\exp\left(-0.2\sqrt{\frac{1}{D}\sum_{i=1}^{D} x_i^2}\right)-\exp\left(\frac{1}{D}\sum_{i=1}^{D}\cos(2\pi x_i)\right)$ | Ackley | MS | 30 | $[-32,32]$ | 0 |
| $f_{c10}(x)=\frac{1}{4000}\sum_{i=1}^{D}(x_i^2)-\left(\prod_{i=1}^{D}\cos\left(\frac{x_i}{\sqrt{i}}\right)\right)+1$ | Griewank | MN | 30 | $[-600,600]$ | 0 |
| $f_{c11}(x)=\frac{\pi}{D}\left\{10\sin^2(\pi y_i)+\sum_{i=1}^{D-1}(y_i-1)^2[1+10\sin^2(\pi y_{i+1})]+(y_D-1)\right\}$ $+\sum_{i=1}^{D} u(x_i, 10, 100, 4)\, y_i=1+\frac{x_i+1}{4}\, u(x_i, a, k, m)=\begin{cases} k(x_i-a)^m & x_i>a \\ 0 & -a<x_i<a \\ k(-x_i-a)^m & x_i<a \end{cases}$ | Penalized | MN | 30 | $[-50,50]$ | 0 |
| $f_{c12}(x)=0.1\{\sin^2(3\pi x_i)+\sum_{i=1}^{D}(x_i-1)^2[1+\sin^2(3\pi x_i)]+(x_D-1)^2[1+\sin^2(2\pi x_D)]\}$ $+\sum_{i=1}^{D} u(x_i, 5, 100, 4)$ | Penalized2 | MN | 30 | $[-50,50]$ | 0 |
| $f_{c13}(x)=\left(\frac{1}{500}+\sum_{j=1}^{25}\frac{1}{j+\sum_{i=1}^{2}[x_j-a_{ij}]^6}\right)^{-1}$ | Foxholes | MS | 2 | $[-65.53, 65.53]$ | 0.998004 |
| $f_{c14}(x)=\sum_{i=1}^{11}\left(a_i-\frac{x_1(b_i^2+b_i x_2)}{b_i^2+b_i x_3+x_4}\right)^2$ | Kowalik | MS | 4 | $[-5,5]$ | 0.0003075 |
| $f_{c15}(x)=4x_1^2-2.1x_1^4+1/3x_1^6+x_1x_2-4x_2^2+x_2^4$ | Six Hump Camel Back | MN | 2 | $[-5,5]$ | −1.03163 |
| $f_{c16}(x)=\left(x_2-\frac{5.1}{4\pi^2}x_1^2+\frac{5}{\pi}x_1-6\right)^2+10\left(1-\frac{1}{8\pi}\right)\cos x_1+10$ | Branin | MS | 2 | $[-5,10]\times[0,15]$ | 0.398 |
| $f_{c17}(x)=[1+(x_1+x_2+1)^2(19-14x_1+3x_1^2-14x_2+6x_1x_2+3x_2^2)]\times[30+(2x_1-3x_2)^2(18-32x_1+12x_1^2+48x_2-36x_1x_2+27x_2^2)]$ | Goldstein Price | MN | 2 | $[-5,5]$ | 3 |
| $f_{c18}(x)=-\sum_{i=1}^{4}(c_i\exp(-\sum_{j=1}^{3} a_{ij}(x_j-p_{ij})^2)$ | Hartman 3 | MN | 3 | $[0,1]$ | −3.8628 |
| $f_{c19}(x)=-\sum_{i=1}^{4}(c_i\exp(-\sum_{j=1}^{6} a_{ij}(x_j-p_{ij})^2)$ | Hartman 6 | MN | 6 | $[0,1]$ | −3.3220 |
| $f_{c20}(x)=-\sum_{i=1}^{5}((X-a_i)(X-a_i)^T+c_i)^{-1}$ | Langermann | MN | 4 | $[0,10]$ | −10.1532 |

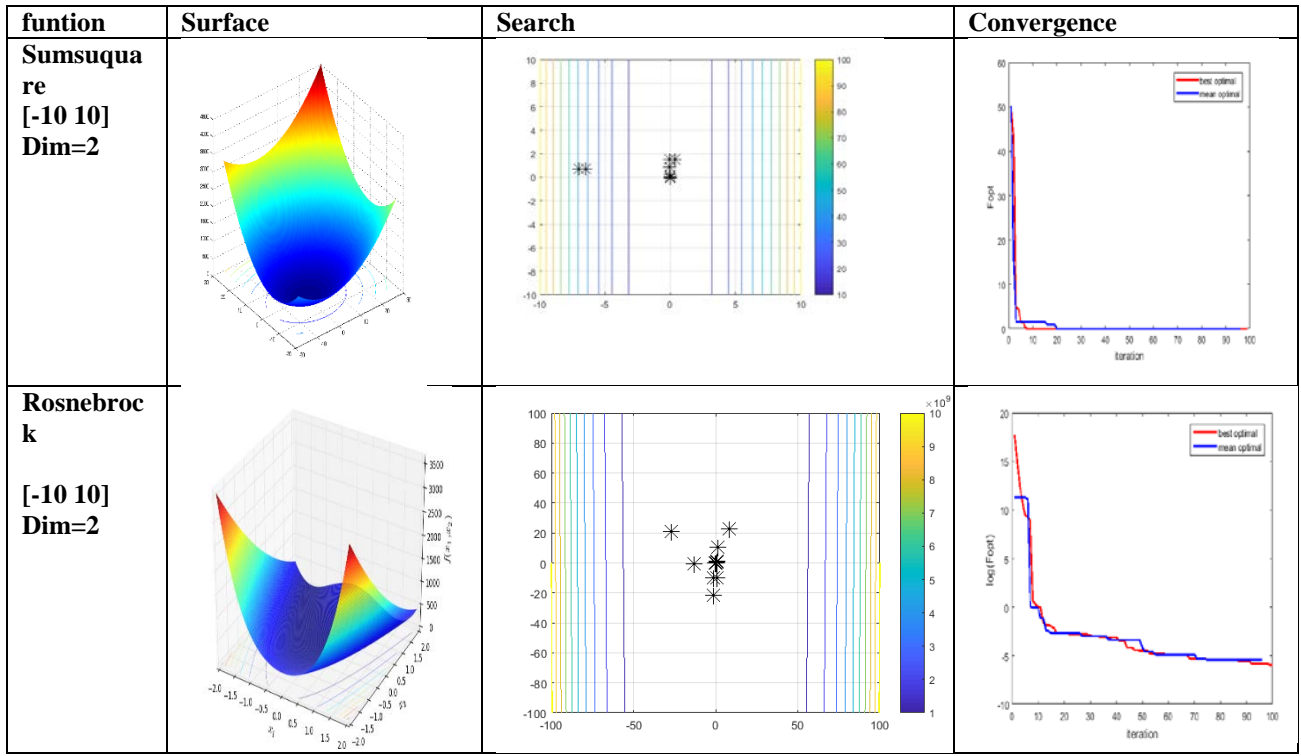

| funtion | Surface | Search | Convergence |
|---|---|---|---|
| **Sumsuquare** [-10 10] Dim=2 | | | |
| **Rosnebrock** [-10 10] Dim=2 | | | |
| **Crossintray** [-10 10] Dim=2 | 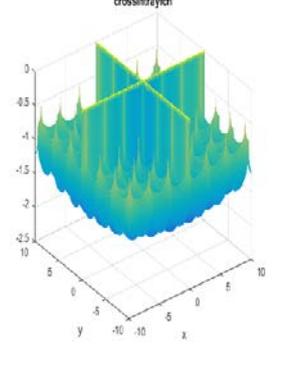 | 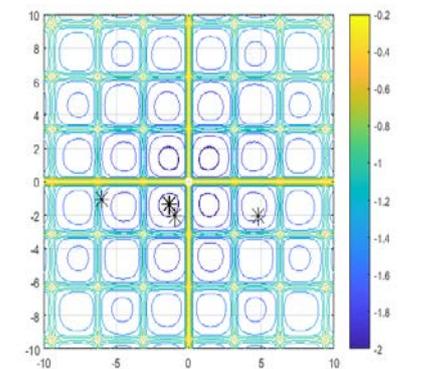 | 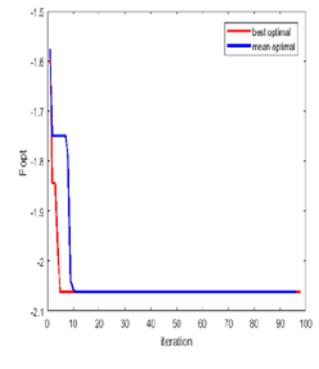 |
| **Ackley** [-35 35] Dim=2 | 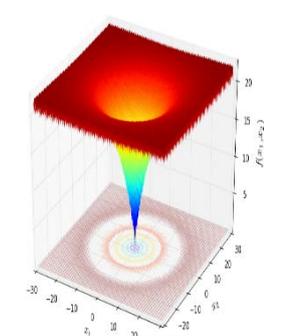 | 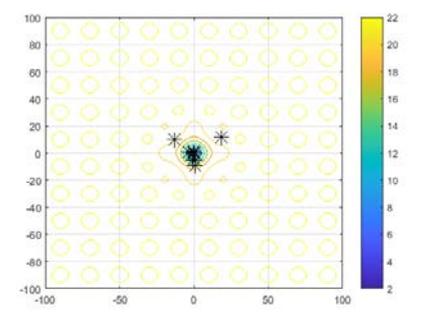 | 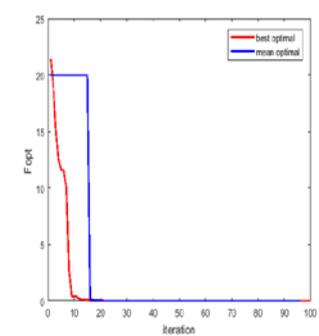 |
| **Shubert** [-5.12 5.12] Dim=2 | 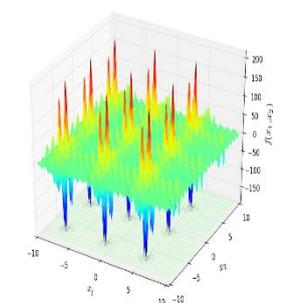 | 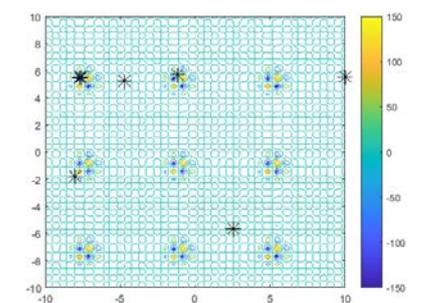 | 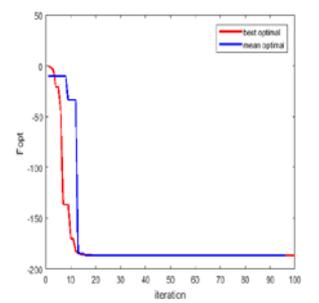 |
| **Rastrigin** [-5.12 5.12] Dim=2 | 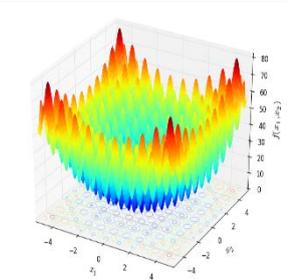 | 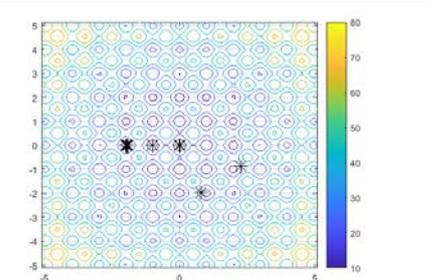 | 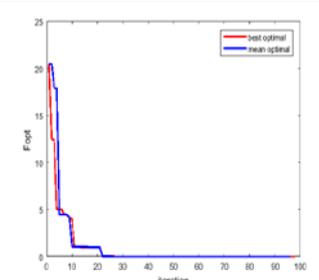 |
| **Foxholes** [-40 40] Dim=2 | 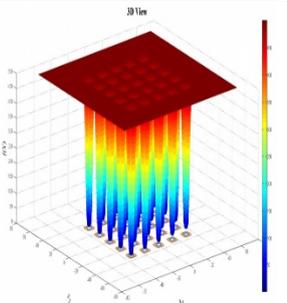 | 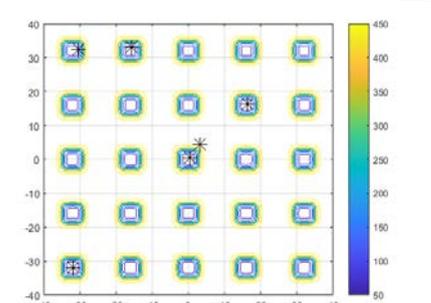 | 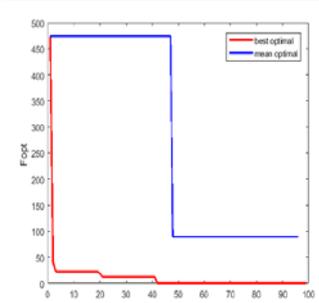 |

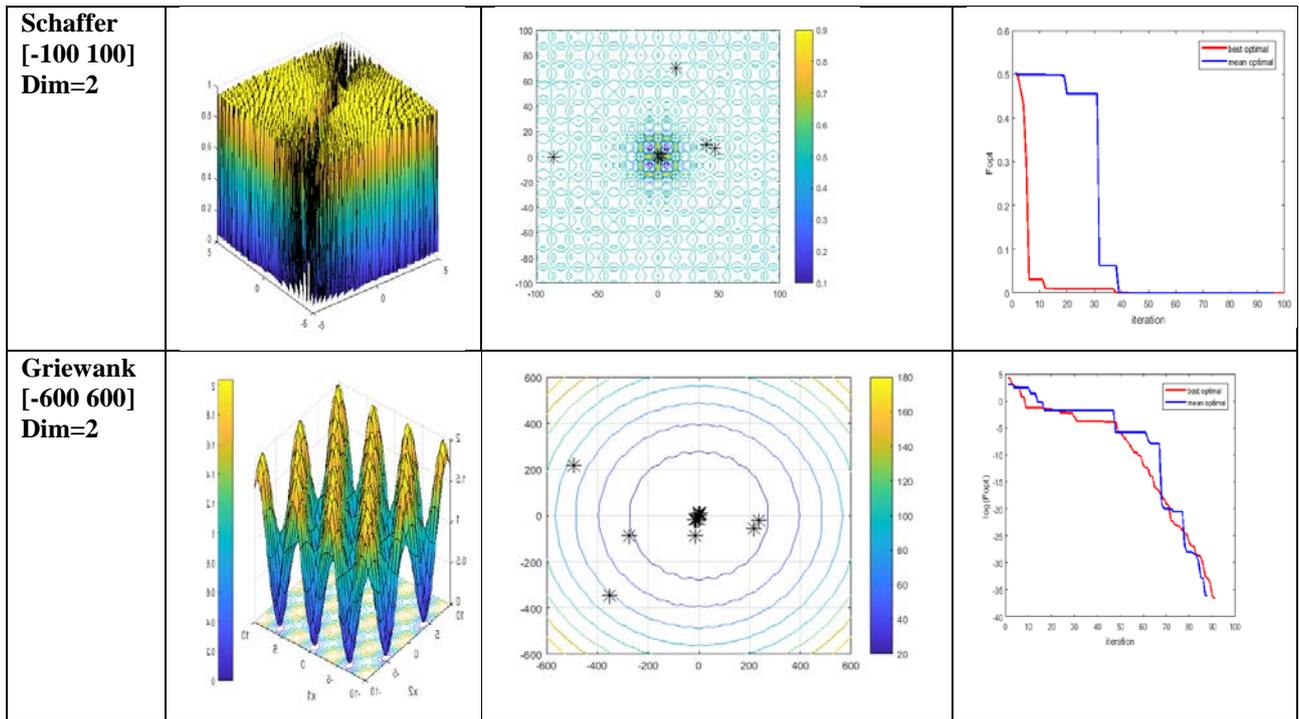

**Figure 6.** Convergence analysis of the proposed TSA and search history on benchmark test problems

**3.2 Experimental results and performance comparison on classical tests**

In this experiment, twenty classical tests are used divided in two classes: unimodal (fc01–fc07) and multimodal (fc08–fc20), the tests (fc13–fc20) are fixed size problem (table 1). The purpose of this experiment is to analyze the feasibility and the success of TSA in classical test functions and to show its exploitation and the exploration search abilities. Beside, TSA is compared to well-known and efficient algorithms. We have chosen in this comparison two algorithms based on trigonometric functions which are Spherical search optimization (SSO) [24] and Sine Cosine Algorithm (SCA) [23]; two powerful algorithms based on evolution strategy which are Evolution Strategy with Covariance Matrix Adaptation (CMA-ES)[32 ], and Success-History Based Parameter Adaptation Differential Evolution (SHADE) (one of the most efficient algorithms in the CEC competition) [33]; we have chosen also one algorithm based swarm optimization ABC [17] , and one recent algorithm called Equilibrium optimizer (EO) [34].

In this experiment, for fair comparison, the population size is set to 20 for all the algorithms except for CMA-ES which requires large population size. The maximum function evaluations (MAX_FE) is set to 50,000 for tests fc01–fc12 and 10,000 for tests fc13–fc20. For each test function, 30 independent runs were carried out. The parameter settings for TSA and the other compared algorithms in this test are detailed as follows:

1) TSA: $Pswitch=0.3$, and the probability of the escape procedure is $Pesc=0.8$
2) SSO: p1=0.5; p2=0.03;
3) SCA: r2∈[0,2*pi], r3∈[0,2];
4) SHADE: F = 0.5, CR = 0.5, n0 = 2, h = 4;
5) ABC: pop_size = 15, Number of Onlooker Bees= 5, Maximum Acceleration= 0.5;
6) EO: a1=2, a2=1, GP=0.5;

The experimental results displayed on the tables 2 and 3 exhibit clearly the potent of TSA to deal with different test functions. In the 30D test functions, TSA shows a clear superiority in both unimodal and multimodal functions. Indeed, the Kruskal-Wallis test (figure 7) confirms this observation. Moreover, the Wilcoxon test shows a clear difference between TSA and the other algorithms used in this experiment (table 4). The nearest algorithm in this experiment is EO, however ABC is not successful in this test which confirms that ABC has slow convergence.

In the fixed size test problems, the results summarized on table 3 display that TSA has a comparative performance with the other algorithms and there is no significant difference between the results of TSA and the other results as confirmed by the Wilcoxon test (table 5). However, in this experiment SSO shows a clear superiority followed by SHADE algorithm (Kruskal-Wallis test figure 8). In this kind of problems, TSA necessities more iterations to converge to the well-known optima.

**Table 2.** Comparative results for 30D test functions

| | | fc01 | fc02 | fc03 | fc04 | fc05 | fc06 | fc07 | fc08 | fc09 | fc10 | fc11 | fc12 |
|---|---|---|---|---|---|---|---|---|---|---|---|---|---|
| **SSO** | mean | 5.43E-11 | 1.95E-05 | 8.36E+03 | 6.67E+00 | 4.10E+01 | **0.00E+00** | 5.57E-02 | 3.02E+01 | 9.63E-05 | 8.50E-08 | 7.01E-08 | 4.15E-11 |
| | std | 2.06E-11 | 6.20E-06 | 2.27E+03 | 1.28E+00 | 2.25E+01 | 0.00E+00 | 1.34E-02 | 3.25E+00 | 4.74E-05 | 9.55E-08 | 5.50E-08 | 2.66E-11 |
| | best | 2.36E-11 | 7.89E-06 | 4.21E+03 | 4.62E+00 | 2.48E+01 | 0.00E+00 | 2.37E-02 | 2.44E+01 | 2.85E-05 | 8.43E-09 | 1.16E-08 | 6.07E-12 |
| **SCA** | mean | 4.27E-08 | 2.14E-12 | 1.72E+03 | 9.11E+00 | 2.85E+01 | **0.00E+00** | 2.05E-02 | 9.45E+00 | 1.34E+01 | 3.74E-02 | 5.08E-01 | 2.25E+00 |
| | std | 1.42E-07 | 7.34E-12 | 3.03E+03 | 8.54E+00 | 1.55E+00 | 0.00E+00 | 2.41E-02 | 2.23E+01 | 9.18E+00 | 1.46E-01 | 1.76E-01 | 1.48E-01 |
| | best | 3.45E-19 | 3.28E-18 | 1.41E+00 | 1.66E-01 | 2.73E+01 | 0.00E+00 | 1.46E-03 | 0.00E+00 | 1.17E-07 | 3.33E-16 | 3.30E-01 | 1.98E+00 |
| **CME-AS** | mean | 8.60E-16 | 3.55E-06 | 3.08E+03 | 7.94E+01 | 7.84E+02 | **0.00E+00** | 9.69E-03 | 1.74E+02 | 1.06E+01 | 1.34E-13 | 3.61E-12 | **3.19E-13** |
| | std | 7.61E-16 | 1.97E-06 | 1.16E+04 | 2.22E+01 | 1.69E+03 | 0.00E+00 | 3.95E-03 | 1.18E+01 | 1.01E+01 | 9.46E-14 | 4.34E-12 | 8.63E-13 |
| | best | 1.67E-16 | 1.20E-06 | 2.90E+01 | 2.53E-05 | 1.57E+01 | 0.00E+00 | 2.40E-03 | 1.36E+02 | 7.25E-09 | 3.20E-14 | 5.94E-14 | 2.95E-15 |
| **SHADE** | mean | 9.46E-25 | 9.41E-16 | 4.76E-04 | 2.73E+01 | **2.49E+01** | 9.00E-01 | 7.24E-02 | 3.32E-02 | 1.62E+00 | 2.00E-02 | **1.12E-26** | 1.04E-01 |
| | std | 5.16E-24 | 4.82E-15 | 7.26E-04 | 6.67E+00 | 2.47E+01 | 3.29E+00 | 4.12E-02 | 1.82E-01 | 1.32E+00 | 2.38E-02 | 3.92E-26 | 2.98E-01 |
| | best | 2.94E-33 | 1.99E-23 | 1.97E-06 | 1.48E+01 | 1.59E-03 | 0.00E+00 | 1.36E-02 | 0.00E+00 | 7.99E-15 | 0.00E+00 | 1.57E-32 | 1.50E-33 |
| **ABC** | mean | 3.73E+01 | 1.27E+00 | 1.01E+04 | 3.80E+01 | 3.50E+03 | 4.72E+01 | 1.38E-01 | 2.21E+02 | 4.24E+00 | 1.22E+00 | 3.93E+01 | 1.20E+01 |
| | std | 9.73E+00 | 2.81E-01 | 1.38E+03 | 3.42E+00 | 1.31E+03 | 8.41E+00 | 3.68E-02 | 1.26E+01 | 4.25E-01 | 6.49E-02 | 6.51E+00 | 2.43E+00 |
| | best | 2.31E+01 | 7.28E-01 | 6.64E+03 | 2.77E+01 | 1.36E+03 | 3.20E+01 | 6.10E-02 | 1.90E+02 | 3.40E+00 | 1.08E+00 | 1.80E+01 | 6.50E+00 |
| **EO** | mean | 1.59E-189 | 1.69E-108 | 4.15E-50 | 2.62E-46 | 2.37E+01 | **0.00E+00** | 3.43E-04 | **0.00E+00** | 4.44E-15 | 0.00E+00 | 5.05E-15 | 7.02E-02 |
| | std | 0.00E+00 | 5.07E-108 | 1.26E-49 | 8.92E-46 | 2.00E-01 | 0.00E+00 | 1.66E-04 | 0.00E+00 | 0.00E+00 | 0.00E+00 | 1.30E-14 | 8.29E-02 |
| | best | 1.50E-199 | 3.85E-112 | 7.66E-63 | 2.68E-53 | 2.32E+01 | 0.00E+00 | 1.56E-05 | 0.00E+00 | 4.44E-15 | 0.00E+00 | 3.39E-19 | 5.26E-16 |
| **TSA** | mean | 0.00E+00 | 2.39E-262 | 0.00E+00 | 2.04E-271 | **1.24E+01** | 0.00E+00 | 1.63E-04 | 0.00E+00 | 8.88E-16 | 0.00E+00 | 3.01E-25 | 1.20E-17 |
| | std | 0.00E+00 | 0.00E+00 | 0.00E+00 | 0.00E+00 | 1.34E+01 | 0.00E+00 | 1.78E-04 | 0.00E+00 | 0.00E+00 | 0.00E+00 | 7.33E-25 | 6.56E-17 |
| | best | 0.00E+00 | 0.00E+00 | 0.00E+00 | 0.00E+00 | 5.17E-08 | 0.00E+00 | 2.03E-06 | 0.00E+00 | 8.88E-16 | 0.00E+00 | 6.47E-29 | 1.41E-27 |

**Table 3.** Comparative results for fixed size test functions

| | | fc13 | fc14 | fc15 | fc16 | fc17 | fc18 | fc19 | fc20 |
|---|---|---|---|---|---|---|---|---|---|
| **SSO** | mean | 1,70E+01 | 1,08E-03 | **-1,03E+00** | 3,98E-01 | **3,00E+00** | -3,86E+00 | **-3,32E+00** | **-1,02E+01** |
| | std | 5,66E+01 | 2,62E-04 | 6,25E-16 | 0,00E+00 | 1,68E-15 | 1,76E-07 | 2,25E-09 | 9,26E-03 |
| | best | 1,00E+00 | 7,51E-04 | -1,03E+00 | 3,98E-01 | 3,00E+00 | -3,86E+00 | -3,32E+00 | -1,02E+01 |
| **SCA** | mean | 2,09E+01 | **1,05E-03** | -1,03E+00 | 4,01E-01 | 3,00E+00 | -3,85E+00 | -2,82E+00 | -1,85E+00 |
| | std | 9,06E+01 | 3,43E-04 | 8,90E-05 | 3,32E-03 | 2,38E-04 | 3,82E-03 | 3,19E-01 | 1,55E+00 |
| | best | 9,98E-01 | 5,30E-04 | -1,03E+00 | 3,98E-01 | 3,00E+00 | -3,86E+00 | -3,15E+00 | -5,43E+00 |
| **CM-EAS** | mean | 4,51E+02 | 4,94E-03 | **-1,03E+00** | 5,42E-01 | 3,78E+00 | -3,84E+00 | -3,22E+00 | -7,83E+00 |
| | std | 1,51E+02 | 5,42E-03 | 6,78E-16 | 5,51E-01 | 4,28E+00 | 1,41E-01 | 1,29E-01 | 3,38E+00 |
| | best | 1,02E+00 | 1,04E-03 | -1,03E+00 | 3,98E-01 | 3,00E+00 | -3,86E+00 | -3,32E+00 | -1,02E+01 |
| **Shade** | mean | 1,78E+01 | 1,68E-03 | **-1,03E+00** | 3,98E-01 | **3,00E+00** | **-3,86E+00** | -3,29E+00 | -8,64E+00 |
| | std | 9,11E+01 | 5,08E-03 | 6,71E-16 | 0,00E+00 | 1,36E-15 | 2,71E-15 | 5,13E-02 | 2,60E+00 |
| | best | 9,98E-01 | 3,07E-04 | -1,03E+00 | 3,98E-01 | 3,00E+00 | -3,86E+00 | -3,32E+00 | -1,02E+01 |
| **ABC** | mean | 5,00E+02 | 1,06E-03 | -1,03E+00 | **3,98E-01** | 3,03E+00 | -3,86E+00 | -3,32E+00 | -9,27E+00 |
| | std | 4,08E-06 | 2,25E-04 | 2,73E-04 | 5,34E-04 | 2,34E-02 | 8,40E-04 | 4,96E-03 | 6,07E-01 |
| | best | 5,00E+02 | 6,48E-04 | -1,03E+00 | 3,98E-01 | 3,00E+00 | -3,86E+00 | -3,32E+00 | -1,00E+01 |
| **EO** | mean | **1,20E+00** | 2,48E-03 | **-1,03E+00** | 3,98E-01 | 5,70E+00 | -3,86E+00 | -3,26E+00 | -7,88E+00 |
| | std | 6,05E-01 | 6,07E-03 | 5,83E-16 | 0,00E+00 | 1,48E+01 | 1,44E-03 | 6,41E-02 | 3,13E+00 |
| | best | 9,98E-01 | 3,07E-04 | -1,03E+00 | 3,98E-01 | 3,00E+00 | -3,86E+00 | -3,32E+00 | -1,02E+01 |
| **TSA** | mean | 5,62E+00 | 3,33E-03 | -1,00E+00 | 3,98E-01 | 1,11E+01 | -3,86E+00 | -3,27E+00 | -1,02E+01 |
| | std | 4,85E+00 | 6,84E-03 | 1,49E-01 | 0,00E+00 | 1,26E+01 | 1,02E-08 | 6,01E-02 | 5,31E-15 |
| | best | 9,98E-01 | 3,08E-04 | -1,03E+00 | 3,98E-01 | 3,00E+00 | -3,86E+00 | -3,32E+00 | -1,02E+01 |

**Table 4.** Wilcoxon test compares TSA to other algorithms on 30d tests (fc01–fc12).

|       | SSO   | SCA   | CMA-ES | SHADE | ABC   | EO    |
|-------|-------|-------|--------|-------|-------|-------|
| P     | 0.000 | 0.000 | 0.000  | 0.000 | 0.000 | 0.003 |
| H     | 1     | 1     | 1      | 1     | 1     | 1     |
| Ranks | +     | +     | +      | +     | +     | +     |

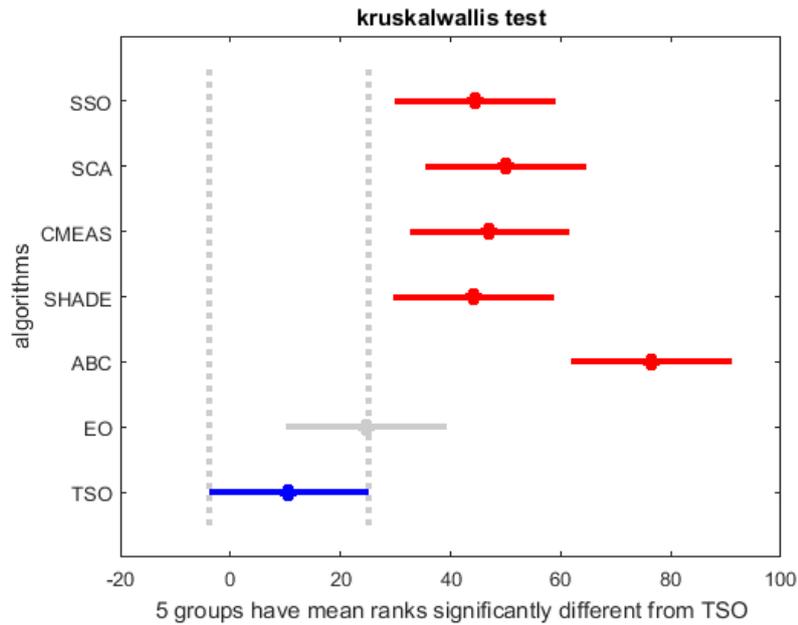

**Figure 7.** Kruskal-Wallis test compare TSA and other algorithms on tests fc01–fc12

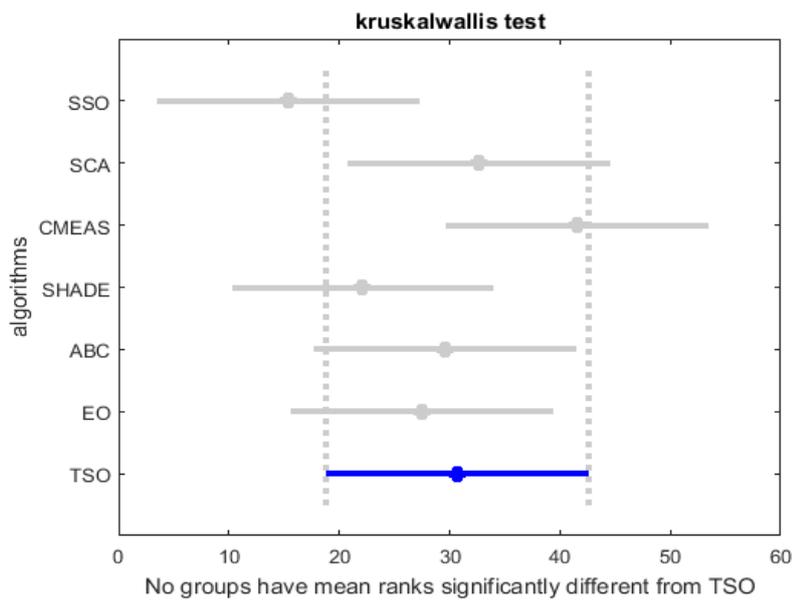

**Figure 8.** Kruskal-Wallis test compares TSA and the other algorithms on tests f13–f20

**Table 5.** Wilcoxon test compares TSA to the other on fixed size tests (f13–f20)

|       | SSO        | SCA        | CMA-ES     | SHADE      | ABC        | EO         |
|-------|------------|------------|------------|------------|------------|------------|
| P     | 2.0313e-01 | 9.4531e-01 | 4.8438e-01 | 2.0313e-01 | 5.4688e-01 | 5.7813e-01 |
| H     | 0          | 0          | 0          | 0          | 0          | 0          |
| ranks | −          | =          | +          | −          | =          | =          |

### 3.3 Experimental results and performance comparison on challenging hard tests

In this experiment, we evaluate the proposed algorithm with top five hard tests for the most algorithms [29], where the overall success is under 3% for many optimization algorithms [29]. Indeed, the overall success of different global optimizers on DeVilliersGlasser02, Damavandi, CrossLegTable, XinSheYang03, and SineEnvelope test function was 0%, 0.25%, 0.83%, 1.08%, and 2.17% respectively [29]. The different information about these hard tests are given in the table 6. In this experiment, we have used the same parameters in the previous experiment, however the maximum function evaluations are fixed to 50,000 for all dimensions. We have compared the results of TSA with those of SSO, SCA, SHADE, and EO.

The results found are summarized in table 7. In the hardest test, DeVilliersGlasser02, the successful algorithm in this test is SHADE algorithm which was able to reach the theoretical global optimum (8,487E-20) followed by TSA algorithm which found a solution near to 0 ( 5,658E-03). However, SSO has the best mean score in this test. In the second hardest problem, Damavandi, our algorithm has 100% overall success. TSA is able to find the theoretical optimum 0 in the 30 runs which confirms the robustness and the effectiveness of the proposed algorithm. The second successful algorithm in this test is SHADE with a best solution equal to 0, however it is trapped in local minima in many runs. In the Crosslegtable problem, TSA, SCA and EO are able to find the best global optimum (-1), however SSO and SHADE algorithms are not successful in this experiment. In the XinSheYang03 test problem, TSA is the only algorithm that reach the optimal global (-1), however, all the other algorithms are trapped in local minima. Finally, in the SineEnvelope test function, TSA has given the nearest best value to the global optimum. SHADE algorithm is the second successful in this test, however SCA is not successful in this problem.

We conclude from this experiment, that TSA has great effectiveness to deal with these hard tests which confirms the exploration capacity of TSA to make a global search and escaping to be tapped in local minima.

**Table 6.** Summary of the hardest tests.

| Test Function | D | Range | Xopt | Fopt |
|---|---|---|---|---|
| $f_{\text{DeVilliersGlasser01}}(\mathbf{x}) = \sum_{i=1}^{24} \left[ x_1 x_2^{t_i} \tanh\left[x_3 t_i + \sin(x_4 t_i)\right] \cos(t_i e^{x_5}) - y_i \right]^2$ Where, $t_i = 0.1(i-1)$ $y_i = 53.81(1.27^{t_i}) \tanh(3.012 t_i + \sin(2.13 t_i)) \cos(e^{0.507} t_i)$ | 5 | [0,60] | Xi=0 | 0 |
| $f_{\text{Damavandi}}(\mathbf{x}) = \left[ 1 - \left\| \frac{\sin[\pi(x_1-2)] \sin[\pi(x2-2)]}{\pi^2(x_1-2)(x_2-2)} \right\|^5 \right] \left[ 2 + (x_1-7)^2 + 2(x_2-7)^2 \right]$ | 2 | [0,14] | Xi=2 | 0 |
| $F_{Crosselegtable}(X)$ $= - \left[ \left\| \sin(x_1) \sin(x_2) e^{\left\| 100 - \sqrt{x_1^2 + x_2^2}/\pi \right\|} \right\| + 1 \right]^{-0.1}$ | 2 | [-10,10] | Xi=0 | -1 |
| $f_{\text{XinSheYang03}}(\mathbf{x}) = e^{-\sum_{i=1}^{n}(x_i/\beta)^{2m}} - 2e^{-\sum_{i=1}^{n} x_i^2} \prod_{i=1}^{n} \cos^2(x_i)$ Where,β=15 and m=3 | 30 | [-20,20] | Xi=0 | -1 |
| $f_{\text{SineEnvelope}}(\mathbf{x}) = - \sum_{i=1}^{n-1} \left[ \frac{\sin^2(\sqrt{x_{i+1}^2 + x_i^2} - 0.5)}{(0.001(x_{i+1}^2 + x_i^2) + 1)^2} + 0.5 \right]$ | 30 | [-100,100] | | −1.49150D + 1.49150 |

**Table7.** statistical results and comparisons for top five hard tests

| Algorithm | | Crosslegtable | XinSheYang03 | Damavandi | DeVilliersGlasser02 | SineEnvelope |
|---|---|---|---|---|---|---|
| | theoretical optimal | -1 | -1 | 0 | 0 | -43.2535 |
| TSA | Best | **-1** | **-1** | **0** | 5.658E-03 | **-43.2534** |
| | Worst | -4.356E-02 | 4.340E-232 | **0.000E+00** | 1.136E+04 | -40.8747 |
| | Mean | -1.409E-01 | **-9.667E-01** | **0.000E+00** | 1.349E+03 | -42.9462 |
| | Std | 1.683E-01 | 1.826E-01 | **0.000E+00** | 2.872E+03 | 0.6241 |
| SSO | Best | -1.362E-03 | 4.340E-232 | 2.000E+00 | 5.120E+00 | -40.5021 |
| | Worst | -2.629E-04 | 4.340E-232 | 2.000E+00 | 3.439E+02 | -37.7725 |
| | Mean | -5.438E-04 | 4.340E-232 | 2.000E+00 | **7.885E+01** | -39.0190 |
| | Std | 2.546E-04 | 0.000E+00 | 0.000E+00 | 7.339E+01 | 0.7357 |
| SCA | Best | **-1** | 4.782E-210 | 4.746E-03 | 5.842E+01 | -35.7815 |
| | Worst | -2.029E-04 | 1.932E-192 | 2.001E+00 | 7.412E+02 | -31.8526 |
| | Mean | **-4.335E-01** | 7.468E-194 | 3.732E-01 | 3.298E+02 | -33.9228 |
| | Std | 5.038E-01 | 0.000E+00 | 7.412E-01 | 1.890E+02 | 1.0578 |
| EO | Best | **-1** | 4.340E-232 | 9.190E-12 | 1.500E+01 | -43.0731 |
| | Worst | -2.558E-03 | 4.340E-232 | 2.000E+00 | 1.047E+04 | -40.8115 |
| | Mean | -1.095E-01 | 4.340E-232 | 7.333E-01 | 1.412E+03 | -42.4431 |
| | Std | 1.695E-01 | 0.000E+00 | 9.803E-01 | 3.131E+03 | 0.5304 |
| SHADE | Best | -8.478E-02 | 4.340E-232 | **0.000E+00** | **8.487E-20** | -43.2510 |
| | Worst | -7.981E-02 | 4.340E-232 | 2.000E+00 | 3.536E+03 | -42.7560 |
| | Mean | -8.422E-02 | 4.340E-232 | 1.933E+00 | 4.472E+02 | **-43.1490** |
| | Std | 1.145E-03 | 0.000E+00 | 3.651E-01 | 8.560E+02 | 0.11782 |

**3.4 Results and comparisons of TSA on CEC 2017 benchmark test functions.**

To more show the effectiveness of the TSA algorithm, the challenging CEC–2017 is used. This benchmark test contains 29 tests composed of shifted and rotated unimodal, multimodal, hybrid, and composition functions [31]. The search domain for all functions are [−100, 100]. In this experiment, each test execution is repeated 30 times and the maximum function evaluation is set to 50,000. We have tested TSA on 10 dimensions and we have used same paramters of the previous experiments. Moreover, TSA results are compared to most popular, recent and high performance algorithms. The comparative methods are EO algorithm, PSO [16], Gravitational Search Algorithm (GSA) [21], Grey Wolf Optimizer (GWO) [35], Salp Swarm Algorithm (SSA) [36], CMA-ES [32], SHADE [33], and SHADE with linear population size reduction hybridized with semi-parameter adaptation of CMA-ES (LSHADE-SPACMA) (one of the CEC 2017 competitors) [37]. The results of the other algorithms are taken from EO paper [34]. The experimental results are summarized in table 8, the best results are in bold.

In this experiment, TSA ranks third after SHADE and LSHADE-SPACM. Statistically, TSA has the same performance as SHADE, the difference between them is not significant as confirmed by the Wilcoxon test displayed in table 9. The remaining algorithms are not successful in this experiment as it is confirmed par Kruskal-Wallis test (figure 9).

**Table 8.** Optimization results and comparison for CEC- 2017 test functions.

| Function | | EO | PSO | GWO | GA | GSA | SSA | CMA-ES | SHADE | LSHADE-SPACMA | TSA |
|---|---|---|---|---|---|---|---|---|---|---|---|
| CEC-2017-f1 | Ave | 2465.3 | 3959.6 | 325 132 | 9799.7 | 296.0 | 3396.25 | **100.00** | **100.00** | **100.00** | 100.00 |
| | Std | 2206.2 | 4456.6 | 107 351 | 5942.54 | 275.1 | 3673.08 | 0.000 | 0.000 | 0.000 | 1.11E-04 |
| CEC-2017-f3 | Ave | 300.00 | 300.00 | 1538.0 | 8721.4 | 10829.2 | **300.00** | 300.00 | 300.00 | 300.00 | 300.00 |
| | Std | 2.4E-08 | 1.9E-10 | 1886.02 | 5900.30 | 1620.74 | 0.00 | 0.000 | 0.000 | 0.000 | 1.41E-06 |
| CEC-2017-f4 | Ave | 404.48 | 405.94 | 409.5 | 410.71 | 406.6 | 406.27 | **400.00** | **400.00** | **400.00** | 400.55 |
| | Std | 0.7911 | 3.28 | 7.55 | 18.512 | 2.92 | 10.07 | 0.000 | 0.000 | 0.000 | 0.54 |
| CEC-2017-f5 | Ave | 510.73 | 513.06 | 513.5 | 516.32 | 556.7 | 521.82 | 530.18 | 503.76 | **502.3** | 517.14 |
| | Std | 3.6707 | 6.54 | 6.10 | 6.926 | 8.40 | 10.50 | 58.32 | 1.006 | 0.87 | 6.25 |
| CEC-2017-f6 | Ave | 600.00 | 600.24 | 600.6 | 600.04 | 621.6 | 609.77 | 682.1 | **600.00** | 600.00 | 600.00 |
| | Std | 1.5E-04 | 0.98 | 0.88 | 0.0668 | 9.015 | 8.26 | 35.43 | 2.6E-07 | 2.59E-07 | 2.12E-03 |
| CEC-2017-f7 | Ave | 720.93 | 718.98 | 729.8 | 728.32 | 714.6 | 740.88 | 713.4 | 713.90 | **711.32** | 725.30 |
| | Std | 5.7425 | 5.10 | 8.60 | 7.290 | 1.55 | 16.62 | 1.63 | 1.23 | 0.37 | 6.55 |
| CEC-2017-f8 | Ave | 809.51 | 811.39 | 814.3 | 820.72 | 820.5 | 823.45 | 828.9 | 803.80 | **801.34** | 817.013 |
| | Std | 2.9176 | 5.47 | 8.26 | 8.961 | 4.69 | 9.95 | 52.98 | 1.27 | 1.03 | 5.80 |
| CEC-2017-f9 | Ave | 900.00 | 900.00 | 911.3 | 910.28 | 900.0 | 944.07 | 4667.3 | **900.00** | **900.00** | 900.06 |
| | Std | 0.0227 | 5.9E-14 | 19.53 | 15.154 | 6.9E-14 | 104.66 | 2062.8 | 0 | 0 | 0.16 |
| CEC-2017-f10 | Ave | 1418.7 | 1473.3 | 1530.5 | 1723.3 | 2694.6 | 1858.85 | 2588.1 | 1193.6 | **1047.2** | 1424.04 |
| | Std | 261.63 | 214.97 | 286.67 | 252.34 | 297.62 | 294.50 | 414.47 | 84.7 | 56.55 | 170.56 |
| CEC-2017-f11 | Ave | 1105.2 | 1110.5 | 1140.2 | 1125.6 | 1134.7 | 1180.5 | 1111.3 | 1100.8 | **1100.0** | 1108.98 |
| | Std | 5.0218 | 6.28 | 54.13 | 23.80 | 10.45 | 59.80 | 25.44 | 1.36 | 4.2E-14 | 4.79 |
| CEC-2017-f12 | Ave | 10 340 | 14 532 | 625 182 | 37 255 | 702 723 | 1 983 166 | 1629.6 | **1324.5** | 1341.7 | 9443.03 |
| | Std | 9790.6 | 11 260 | 1 126 443 | 34792.7 | 42075.4 | 1 909 901 | 198.11 | 102.6 | 86.25 | 13478.90 |
| CEC-2017-f13 | Ave | 8023.0 | 8601.1 | 9842.3 | 10 828 | 11 053 | 16098.6 | 1323.6 | 1304.7 | **1303.7** | 1319.65 |
| | Std | 6720.8 | 5123.6 | 5633.43 | 8928.94 | 2110.55 | 10537.2 | 78.32 | 0.71 | 3.25 | 10.74 |
| CEC-2017-f14 | Ave | 1463.3 | 1482.1 | 3403.53 | 7048.9 | 7147.5 | 1508.94 | 1452.1 | 1410.8 | **1400.2** | 1415.47 |
| | Std | 32.498 | 42.46 | 1953.33 | 8160.08 | 1489.52 | 51.05 | 55.98 | 9.21 | 0.44 | 10.93 |
| CEC-2017-f15 | Ave | 1585.6 | 1714.3 | 3806.60 | 9296.2 | 18 001 | 2236.69 | 1509.6 | 1500.3 | **1500.3** | 1504.34 |
| | Std | 48.012 | 282.89 | 3860.66 | 8978.18 | 5498.67 | 571.19 | 16.43 | 0.36 | 0.20 | 5.20 |
| CEC-2017-f16 | Ave | 1649.0 | 1860.0 | 1725.78 | 1786.3 | 2149.7 | 1726.26 | 1815.3 | 1602.5 | **1600.9** | 1645.92 |
| | Std | 50.915 | 127.65 | 123.85 | 129.07 | 105.8 | 126.97 | 230.1 | 2.19 | 0.36 | 59.37 |
| CEC-2017-f17 | Ave | 1731.6 | 1761.6 | 1759.61 | 1746.5 | 1857.7 | 1774.57 | 1830.1 | 1716.4 | **1700.4** | 1709.62 |
| | Std | 18.071 | 47.50 | 31.29 | 39.78 | 108.32 | 34.23 | 175.8 | 5.96 | 0.35 | 8.18 |
| CEC-2017-f18 | Ave | 12 450 | 14 599 | 25806.1 | 15 721 | 8720.5 | 23429.1 | 1825.9 | 1809.9 | **1801.1** | 1820.95 |
| | Std | 11 405 | 11852.2 | 15766.9 | 12 828 | 5060.1 | 14045.7 | 13.53 | 9.46 | 3.67 | 9.82 |
| CEC-2017-f19 | Ave | 1951.5 | 2602.8 | 9866.1 | 9686.5 | 13 670 | 2916.1 | 1920.5 | 1900.5 | **1900.3** | 1901.93 |
| | Std | 47.108 | 2185.02 | 6371.09 | 6766.3 | 19 168 | 1871.2 | 28.68 | 0.28 | 0.39 | 1.51 |
| CEC-2017-f20 | Ave | 2020.6 | 2085.1 | 2075.6 | 2056.5 | 2272.3 | 2089.3 | 2494.8 | 2020.0 | **2000.2** | 2005.03 |
| | Std | 22.283 | 62.25 | 52.04 | 60.01 | 81.72 | 49.28 | 242.65 | 0.0093 | 0.14 | 7.76 |
| CEC-2017-f21 | Ave | 2307.5 | 2281.7 | 2317.1 | 2303.8 | 2357.7 | 2249.8 | 2324.7 | 2282.5 | **2202.1** | 2227.83 |
| | Std | 20.961 | 54.02 | 7.00 | 43.75 | 28.20 | 60.44 | 67.76 | 42.6 | 4.02 | 48.90 |
| CEC-2017-f22 | Ave | 2297.4 | 2314.8 | 2310.1 | 2304.6 | 2300.0 | 2301.5 | 3532.4 | 2297.3 | 2300.1 | **2291.64** |
| | Std | 18.402 | 66.10 | 16.75 | 2.38 | 0.072 | 11.80 | 847.6 | 16.18 | 0.17 | 24.30 |
| CEC-2017-f23 | Ave | 2615.8 | 2620.8 | 2616.4 | 2632.9 | 2736.5 | 2621.7 | 2728.8 | 2608.1 | **2603.3** | 2612.27 |
| | Std | 5.5298 | 9.23 | 8.47 | 13.42 | 39.14 | 8.69 | 243.1 | 1.71 | 1.41 | 3.96 |
| CEC-2017-f24 | Ave | 2743.8 | 2692.2 | 2741.7 | 2758.3 | 2742.2 | 2733.2 | 2704.4 | 2728.9 | **2677.9** | 2728.45 |
| | Std | 6.904 | 108.20 | 8.73 | 14.92 | 5.52 | 64.43 | 73.42 | 31.7 | 91.6 | 62.37 |
| CEC-2017-f25 | Ave | 2934.3 | 2924.0 | 2938.0 | 2947.9 | 2937.5 | 2923.5 | 2932.01 | 2916.4 | 2930.0 | **2881.46** |
| | Std | 19.76 | 25.02 | 23.61 | 19.25 | 15.36 | 23.86 | 20.87 | 22.9 | 21.3 | 77.36 |
| CEC-2017-f26 | Ave | 2967.8 | 2952.1 | 3222.5 | 3112.1 | 34407.5 | 2900.9 | 3457.7 | 2909.2 | 2900.0 | **2854.91** |
| | Std | 164.98 | 249.66 | 427.02 | 334.65 | 628.73 | 36.56 | 598.9 | 34.9 | 0 | 105.25 |
| CEC-2017-f27 | Ave | 3091.3 | 3116.2 | 3104.1 | 3115.1 | 3259.5 | 3092.6 | 3137.5 | 3071.5 | **3089.5** | 3090.90 |
| | Std | 2.2414 | 24.99 | 21.81 | 19.18 | 41.66 | 2.78 | 21.37 | 0.78 | 0.15 | 1.95 |
| CEC-2017-f28 | Ave | 3302.7 | 3315.9 | 3391.2 | 3320.7 | 3459.4 | 3210.5 | 3397.6 | 3266.7 | **3125.0** | 3191.13 |
| | Std | 133.92 | 121.83 | 101.5 | 126.34 | 33.84 | 113.17 | 131.3 | 22.2 | 63.2 | 162.20 |
| CEC-2017-f29 | Ave | 3169.9 | 3203.8 | 3190.5 | 3253.5 | 3449.5 | 3214.1 | 3213.5 | 3142.8 | **3134.9** | 3154.65 |
| | Std | 24.65 | 52.26 | 42.9 | 81.99 | 171.33 | 51.69 | 109.79 | 12.9 | 3.87 | 16.812 |
| CEC-2017-f30 | Ave | 297 113 | 350 650 | 297688 | 537277 | 1 303 361 | 421 120 | 304 569 | 3201.1 | **3430.6** | 47261.20 |
| | Std | 458560 | 504857 | 527757 | 637410 | 363843 | 568 085 | 444 815 | 0.31 | 0.31 | 227528.47 |

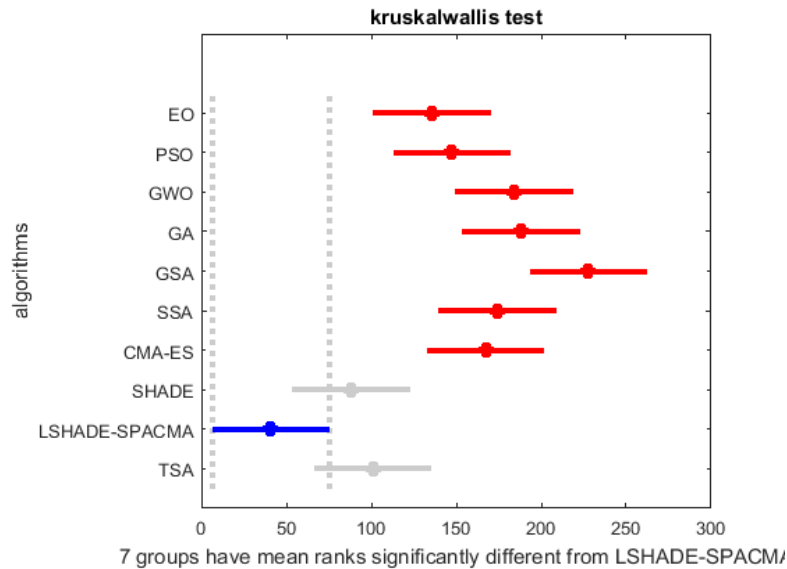

**Figure 9.** Kruskal Wallis test compares TSA and the other algorithms in 10d CEC 2017 test functions

**Table.9.** Wilcoxon test for 10d CEC 2017 test function

|        | EO     | PSO    | GWO    | GA     | GSA    | SSA    | CMA-ES | SHADE  | LSHADE-SPACMA |
|--------|--------|--------|--------|--------|--------|--------|--------|--------|---------------|
| Pvalue | 0.0057 | 0.0031 | 0.0000 | 0.0000 | 0.0000 | 0.0000 | 0.0005 | 0.1742 | 0.0006        |
| h      | 1      | 1      | 1      | 1      | 1      | 1      | 1      | 0      | 1             |
| ranks  | +      | +      | +      | +      | +      | +      | +      | =      | −             |

## 4. Conclusion

In this article, a novel optimization algorithm called Tangent Search Algorithm for optimization probelms (TSA) is presented. The TSA is based on new flight equation based on mathematical tangent function. To provide a better balance between the explorative and exploitative search capacities, the proposed algorithm is composed of three components: exploration search, exploration search, and escape local procedure. and local search phase. Besides, the use of the variable step-size helps to reduce the magnitude of movement of the solutions in the last iterations. The efficacy of TSA is validated throughout quantitative and qualitative metrics by testing it on classical tests, five challenging tests, the CEC2017, the most recent CEC2020, and four constraint engineering problems. Moreover, comparison with several popular, recent, and high-performance optimization algorithms displayed a high effectiveness of the proposed TSA algorithm in locating the optimal or near-optimal solutions with higher efficiency. Future studies should work toward developing binary and multi-objective varieties of the TSA algorithm and to incorporate some tricks to more improve its performance.

**Declaration of conflict of interest**

The authors have no conflicts of interest to declare that are relevant to the content of this article.